\pdfoutput=1

\documentclass[11pt]{article}

\usepackage{acl}

\usepackage{amsmath} 
\usepackage{graphicx}
\usepackage{hyperref}
\usepackage[capitalize]{cleveref}
\Crefformat{figure}{#2Fig.~#1#3}
\Crefmultiformat{figure}{Figs.~#2#1#3}{ and~#2#1#3}{, #2#1#3}{ and~#2#1#3}

\usepackage{times}
\usepackage{latexsym}

\usepackage[T1]{fontenc}

\usepackage[utf8]{inputenc}

\usepackage{microtype}

\usepackage{xcolor}

\usepackage[normalem]{ulem} 

\title{Team \'{U}FAL at CMCL 2022 Shared Task: Figuring out the correct recipe for predicting Eye-Tracking features using Pretrained Language Models}

\author{Sunit Bhattacharya, Rishu Kumar \and Ond\v{r}ej Bojar \\
        Charles University \\ 
        Faculty Of Mathematics and Physics \\
        Insititute of Formal and Applied Linguistics \\
        \texttt{{bhattacharya,kumar,bojar}@ufal.mff.cuni.cz} \\
  }

\begin{document}
\maketitle

\begin{abstract}
Eye-Tracking data is a very useful source of information to study cognition and especially language comprehension in humans. In this paper, we describe our systems for the CMCL 2022 shared task on predicting eye-tracking information. We describe our experiments with pretrained models like BERT and XLM and the different ways in which we used those representations to predict four eye-tracking features. Along with analysing the effect of using two different kinds of pretrained multilingual language models and different ways of pooling the token-level representations, we also explore how contextual information affects the performance of the systems. Finally, we also explore if factors like augmenting linguistic information affect the predictions. Our submissions achieved
an average MAE of 5.72 and ranked $5^{th}$ in the shared task. The average MAE showed further reduction to 5.25 in post task evaluation.
\end{abstract}

\section{Introduction and Motivation}
\label{intro}
In the last decade that has seen rapid developments in AI research, the emergence of the Transformer architecture \cite{vaswani2017attention} marked a pivotal point in Natural Language Processing (NLP). Fine-tuning pretrained language models to work on various downstream tasks has become a dominant method of obtaining state-of-the-art performance in different areas. Their capability to capture linguistic knowledge and learn powerful contextual word embeddings \cite{liu2019linguistic} have made the transformer based models the work-horses in many NLP tasks. Pretrained models like the multilingual BERT \cite{devlin2019bert} and XLM \cite{conneau2020unsupervised} have also shown state-of-the-art performance on cross-lingual understanding tasks \cite{wu-dredze-2019-beto,artetxe2019cross}. In some cases like machine translation, there are even claims that deep learning systems reach translation qualities that are comparable to professional translators \cite{popel2020transforming}.

Language processing and its links with cognition is a very old research problem which has revealed how cognitive data (eg. gaze, fMRI) can be used to investigate human cognition. Attempts at using computational methods for such studies \cite{mitchell2008predicting,dehghani2017decoding} have also shown encouraging results. However recently, there have been a number of works that have tried to incorporate human cognitive data collected during reading for improving the performance of NLP systems \cite{hollenstein2019advancing}. The CMCL 2022 Shared Task of multilingual and cross-lingual prediction of human reading behavior \cite{hollenstein2022shared} explores how eye-gaze attributes can be algorithmically predicted given reading data in multilingual settings. 

Informed by the previous attempts at using pretrained multilingual language models to predict human reading behavior \cite{hollenstein-etal-2021-multilingual} we experiment with multilingual BERT and XLM based models to test which fares better in this task. For the experiments with the pretrained models, we use the trained weights from Huggingface~\cite{wolf-etal-2020-transformers} and perform the rest of our experiments using PyTorch\footnote{https://pytorch.org/}.
Inspired by the psycholinguistic research on investigating context length during processing \cite{wochna2013context}, we experiment how different contexts affect model performance. Finally, we merged the principles of the "classical" approach of feature-based prediction with the pretrained-language model based prediction for further analysis. In the following sections, we present our results from a total of 48 different models. 

\section{Task Description}
\label{taskdescription}
The CMCL 2022 Shared Task of Multilingual and Cross-lingual prediction of human reading behavior frames the task of predicting eye-gaze attributes associated with reading sentences as a regression task. The data for the task was comprised of eye movements corresponding to reading sentences in six languages (Chinese, Dutch, English, German, Hindi, Russian). The training data for the task contained 1703 sentences while the development set and test set contained 104 and 324 sentences respectively. The data was presented in a way such that for each word in a sentence there were four associated eye-tracking features in the form of the mean and standard deviation scores of the Total Reading Time (TRT) and First Fixation Duration (FFD). The features in the data were scaled in the range between 0 and 100 to facilitate evaluation via the mean absolute average (MAE).


\section{Experiments}
A total of 48 models of different configurations were trained with the data provided for the shared task. The different configurations used to construct the models are based on intuition and literature survey.

Thee models were primarily categorized as System-1 (sys1) and System-2 (sys2) models. For some word corresponding to a sentence in the dataset, System-1 models provided no additional context information. System-2 models on the other hand, contained the information of all the words in the sentence that preceded the current word, providing additional context. This setting was inspired by works \cite{khandelwal2018sharp,clark2019does} on how context is used by language models. 

All systems under the System-1/2 labels were further trained as a BERT (bert) based system or a XLM (xlm) based system. BERT embeddings were previously used by  \citet{choudhary2021mtl782_iitd} for the eye-tracking feature prediction task in CMCL 2021.

Corresponding to each such language models (bert and xlm), the impact of different fine-tuning strategies\cite{sun2019fine} on system performance was studied. Hence, for one setting, only the contextualized word representation (CWR) was utilized by freezing the model weights and putting a learnable regression layer on top of the model output layer (classifier). Alternatively, the models were fine-tuned with the regression layer on top of them (whole). This setting is similar to the one used by \citet{li2021torontocl}. However in our case, we experiment with a BERT and XLM pretrained model. 

Additionally, we also performed experiments with pooling strategies for the layer representations by either using the final hidden representation of the first sub-word encoding of the input (first) or aggregating the representations of all sub-words using mean-pooling (mean) or sum-pooling (sum). The rationale behind using different pooling strategies was to have a sentence-level representation of the input tokens. The impact of different pooling strategies has previously been studied \cite{shao2019transformer,lee2019set} for different problems. In this paper, we analyze the effect of pooling feature-space embeddings in the context of eye-tracking feature prediction.       

Finally, for the experiments where we augmented additional lexical features (augmented) to the neural features for regression, we used word length and word-frequency as the additional information following \citet{vickers-etal-2021-cognlp}.

Constructing the experiments in this manner provided us with models with a diverse set of properties and in turn provided insights into how well the model behaves when all other things stay the same, and only one aspect of learning is changed.

\section{Results}
The results corresponding to the top 10 systems based on the experiments described above are shown in \cref{table:1}. 

\begin{table}[h!]
\centering
\begin{tabular}{|c|c|} 
\hline
Model & MAE \\
\hline
bert\_sys2\_augmented\_sum\_classifier & 5.251 \\
\hline
bert\_sys2\_unaugmented\_first\_classifier & 5.267 \\
\hline
bert\_sys2\_augmented\_mean\_classifier & 5.272 \\
\hline
bert\_sys1\_augmented\_mean\_classifier & 5.279 \\
\hline
bert\_sys2\_augmented\_first\_classifier & 5.295 \\
\hline
xlm\_sys1\_augmented\_first\_classifier & 5.341 \\
\hline
xlm\_sys2\_augmented\_first\_whole & 5.346 \\
\hline
bert\_sys1\_augmented\_sum\_classifier & 5.353 \\
\hline
bert\_sys2\_augmented\_sum\_whole & 5.367\\
\hline
xlm\_sys2\_augmented\_first\_classifier & 5.373 \\
\hline
\end{tabular}
\caption{Top 10 best performing systems}
\label{table:1}
\end{table}

It was observed that the maximum MAE scores (and the maximum variance of scores)  for all the models was obtained for the attribute "TRT\_Avg". The attribute wise variances corresponding to the test-data for all the models are shown in \cref{table:2}. Similarly, the mean values of the attributes for all models are shown in \cref{table:3}.

\begin{table}[h!]
\centering
\begin{tabular}{|c|c|c|c|} 
\hline
FFD\_Avg & FFD\_Std & TRT\_Avg & TRT\_Std  \\
\hline
0.194 & 0.403 & 0.637 & 0.489\\
\hline
\end{tabular}
\caption{Attribute wise variance of scores for all models}
\label{table:2}
\end{table}

\begin{table}[h!]
\centering
\begin{tabular}{|c|c|c|c|} 
\hline
FFD\_Avg & FFD\_Std & TRT\_Avg & TRT\_Std  \\
\hline
5.691 & 2.646 & 8.633 &	5.806\\
\hline
\end{tabular}
\caption{Attribute wise mean of scores for all models}
\label{table:3}
\end{table}

An analysis of the models based on the different experimental configurations are described in the following sections.

\subsection{System-1 vs System-2}
\cref{table:12} shows the average model performance across System-1 and System-2 configurations for both BERT and XLM based models (based on the average MAE values of the configurations). We see that for the BERT based models, the average MAE for System-1 is lower than that of System-2. But for XLM-based models, the difference is almost non-existent. 

\begin{table}[h!]
\centering
\begin{tabular}{|c|c|} 
\hline
Model & Average MAE across models  \\
\hline
Sys1\_BERT & 5.66 \\
\hline
Sys1\_XLM & 5.70 \\
\hline
Sys2\_BERT & 5.72 \\
\hline
Sys2\_XLM & 5.69 \\
\hline
\end{tabular}
\caption{System-1 vs System-2 performance across models}
\label{table:12}
\end{table}

However, it should be noted that 12 out of the first 20 best performing models were System-2 models. Hence we posit that although the availability of the full sentence context is a factor for having more efficient systems, independently the factor does not seem to boost the overall performance much.

\subsection{BERT vs XLM}
\cref{table:13} shows that there is only a tiny difference in average MAE for all four attributes (FFD\_$\mu$, FFD\_$\sigma$, TRT\_$\mu$, TRT\_$\sigma$) for all BERT vs XLM models . However, a brief look at \cref{table:4} and \cref{table:5} reveal that it was the XLM models that were responsible for slightly decreased MAE scores for 3 of the 4 attributes that were being predicted.

\begin{table}[h!]
\centering
\begin{tabular}{|c|c|} 
\hline
Model & Average MAE across models  \\
\hline
BERT & 5.6920 \\
\hline
XLM & 5.6960 \\
\hline
\end{tabular}
\caption{BERT vs XLM performance across models}
\label{table:13}
\end{table}

\begin{table}[h!]
\centering
\begin{tabular}{|c|c|c|c|c|} 
\hline
Model & FFD\_$\mu$ & FFD\_$\sigma$ & TRT\_$\mu$ & TRT\_$\sigma$  \\
\hline
BERT & 0.141 & 0.776 & 0.952 & 0.792\\
\hline
XLM & 0.236 & 0.045 & 0.349 & 0.204 \\
\hline
\end{tabular}
\caption{Attribute wise variance of scores for all BERT and XLM based models}
\label{table:4}
\end{table}

\begin{table}[h!]
\centering
\begin{tabular}{|c|c|c|c|c|} 
\hline
Model & FFD\_$\mu$ & FFD\_$\sigma$ & TRT\_$\mu$ & TRT\_$\sigma$ \\
\hline
BERT & 5.592 & 2.679 & 8.645 & 5.852\\
\hline
XLM & 5.789 & 2.612 & 8.622 & 5.760\\
\hline
\end{tabular}
\caption{Attribute wise mean of scores for all BERT and XLM based models}
\label{table:5}
\end{table}

We also see that the amount of variance for XLM based models was also smaller for 3 of the 4 attributes. 

\subsection{Augmented vs Un-Augmented models}

\begin{figure}[h!]
    \centering
    \includegraphics[width=7cm]{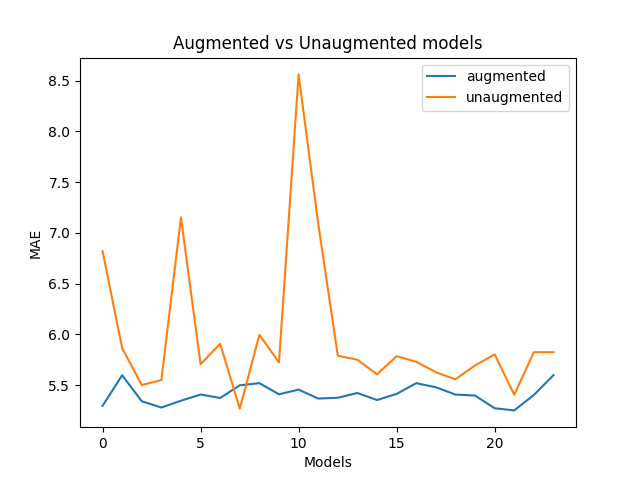}
    \caption{Augmented vs Un-augmented model performance. The x-axis represents the 24 different models of each category. The y-axis shows the MAE corresponding to each model.}
    \label{fig:aug_uaug}
\end{figure}

\cref{fig:aug_uaug} shows that augmented models. i.e. models that were fed information like word-frequency and word-length along with the neural representation information before being fed to the regression layer performed better than models that used only contextual word embeddings resulting from pretrained language models. \cref{table:14} and \cref{table:15} show the 5 best performing models of this category sorted by their MAE. 

\begin{table}[h!]
\centering
\begin{tabular}{|c|c|} 
\hline
Model & MAE \\
\hline
bert\_sys2\_unaugmented\_first\_classifier & 5.267\\
\hline
bert\_sys2\_unaugmented\_mean\_classifier & 5.405\\
\hline
xlm\_sys1\_unaugmented\_mean\_classifier & 5.5\\
\hline
xlm\_sys2\_unaugmented\_mean\_classifier & 5.55\\
\hline
xlm\_sys1\_unaugmented\_mean\_classifier & 5.557 \\
\hline
\end{tabular}
\caption{Performance of 5 best Un-Augmented models.}
\label{table:14}
\end{table}

\begin{table}[h!]
\centering
\begin{tabular}{|c|c|} 
\hline
Model & MAE \\
\hline
bert\_sys2\_augmented\_sum\_classifier&5.251\\
\hline
bert\_sys2\_augmented\_mean\_classifier&5.272\\
\hline
bert\_sys1\_augmented\_mean\_classifier&5.279\\
\hline
bert\_sys2\_augmented\_first\_classifier&5.295\\
\hline
xlm\_sys1\_augmented\_first\_classifier&5.341\\
\hline
\end{tabular}
\caption{Performance of 5 best Augmented models}
\label{table:15}
\end{table}

\begin{table}[h!]
\centering
\begin{tabular}{|c|c|c|c|c|} 
\hline
Model & FFD\_$\mu$ & FFD\_$\sigma$ & TRT\_$\mu$ & TRT\_$\sigma$  \\
\hline
Aug  & 5.502 & 2.511 & 8.181 & 5.436 \\
\hline
Uaug  & 5.88 & 2.78 & 9.086 & 6.176\\
\hline
\end{tabular}
\caption{Attribute wise mean of scores for all Augmented and Un-augmented models}
\label{table:6}
\end{table}

\begin{table}[h!]
\centering
\begin{tabular}{|c|c|c|c|c|} 
\hline
Model & FFD\_$\mu$ & FFD\_$\sigma$ & TRT\_$\mu$ & TRT\_$\sigma$ \\
\hline
Aug  & 0.017 & 0.004 & 0.015 & 0.007 \\
\hline
Uaug  & 0.292 & 0.749 & 0.823 & 0.678 \\
\hline
\end{tabular}
\caption{Attribute wise variance of scores for all Augmented and Un-augmented models}
\label{table:7}
\end{table}

The mean and variance of attributes across models of these families presented in \cref{table:6} \& \ref{table:7} show that augmented models show way less variance in their predictions in comparison with neural-representation only model families.

\subsection{Nature of representation of input tokens (Pooling strategies)}

\begin{figure}[h!]
    \centering
    \includegraphics[width=7cm]{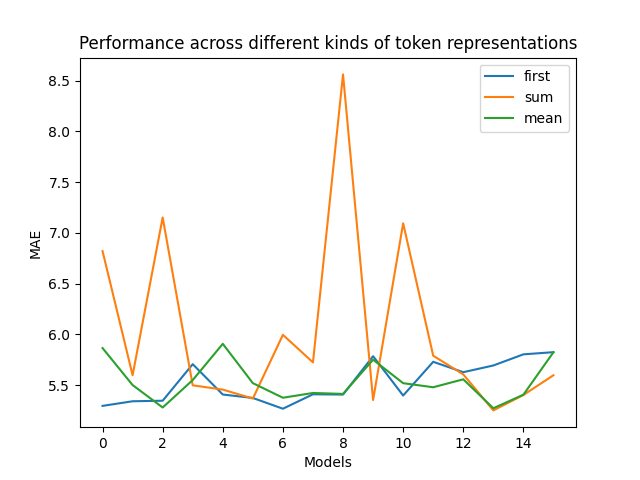}
    \caption{Model performance based on the nature of representation of input tokens.The x-axis represents the 16 different models of each category. The y-axis shows the MAE corresponding to each model.}
    \label{fig:cls_mean_sum}
\end{figure}

\cref{fig:cls_mean_sum} shows that using the first sub-word token or the mean-pooled representation of the entire input gives lesser MAE scores than the sum-pooled representations. It was also observed that for System-2 family of models, the mean-pooled representations were associated with lesser MAE scores in comparison to the first sub-word representation. The attribute wise mean in \cref{table:8} and attribute wise variance of model MAEs shown in \cref{table:9} illustrates this point. \cref{table:16},\cref{table:17} and \cref{table:18} show the 5 best performing models of this category sorted by their MAE.

\begin{table}[h!]
\centering
\begin{tabular}{|c|c|} 
\hline
Model & MAE \\
\hline
bert\_sys2\_unaugmented\_first\_classifier&5.267\\
\hline
bert\_sys2\_augmented\_first\_classifier&5.295\\
\hline
xlm\_sys1\_augmented\_first\_classifier&5.341\\
\hline
xlm\_sys2\_augmented\_first\_whole&5.346\\
\hline
xlm\_sys2\_augmented\_first\_classifier&5.373\\
\hline
\end{tabular}
\caption{Performance of 5 best first models}
\label{table:16}
\end{table}

\begin{table}[h!]
\centering
\begin{tabular}{|c|c|} 
\hline
Model & MAE \\
\hline
bert\_sys2\_augmented\_mean\_classifier&5.272\\
\hline
bert\_sys1\_augmented\_mean\_classifier&5.279\\
\hline
bert\_sys2\_augmented\_mean\_whole&5.375\\
\hline
bert\_sys2\_unaugmented\_mean\_classifier&5.405\\
\hline
xlm\_sys1\_augmented\_mean\_whole&5.413\\
\hline
\end{tabular}
\caption{Performance of 5 best Mean models}
\label{table:17}
\end{table}

\begin{table}[h!]
\centering
\begin{tabular}{|c|c|} 
\hline
Model & MAE \\
\hline
bert\_sys2\_augmented\_sum\_classifier&5.251\\
\hline
bert\_sys1\_augmented\_sum\_classifier&5.353\\
\hline
bert\_sys2\_augmented\_sum\_whole&5.367\\
\hline
bert\_sys1\_augmented\_sum\_whole&5.402\\
\hline
xlm\_sys2\_augmented\_sum\_classifier&5.456\\
\hline
\end{tabular}
\caption{Performance of 5 best Sum models}
\label{table:18}
\end{table}

\begin{table}[h!]
\centering
\begin{tabular}{|c|c|c|c|c|} 
\hline
Model & FFD\_$\mu$ & FFD\_$\sigma$ & TRT\_$\mu$ & TRT\_$\sigma$  \\
\hline
first  & 5.549 & 2.505 & 8.434 & 5.615 \\
\hline
Mean  & 5.57 & 2.538 & 8.416 & 5.636 \\
\hline
Sum  & 5.954 & 2.894 & 9.05 & 6.167 \\
\hline
\end{tabular}
\caption{Attribute wise mean of scores for models with different input token representations}
\label{table:8}
\end{table}

\begin{table}[h!]
\centering
\begin{tabular}{|c|c|c|c|c|} 
\hline
Model & FFD\_$\mu$ & FFD\_$\sigma$ & TRT\_$\mu$ & TRT\_$\sigma$ \\
\hline
first  & 0.036 & 0.004 & 0.118 & 0.054 \\
\hline
Mean  & 0.047 & 0.005 & 0.118 & 0.048 \\
\hline
Sum  & 0.383 & 1.082 & 1.374 & 1.139 \\
\hline
\end{tabular}
\caption{Attribute wise variance of scores for models with different input token representations}
\label{table:9}
\end{table}

\subsection{Fine-tuning}

\begin{figure}[h!]
    \centering
    \includegraphics[width=7cm]{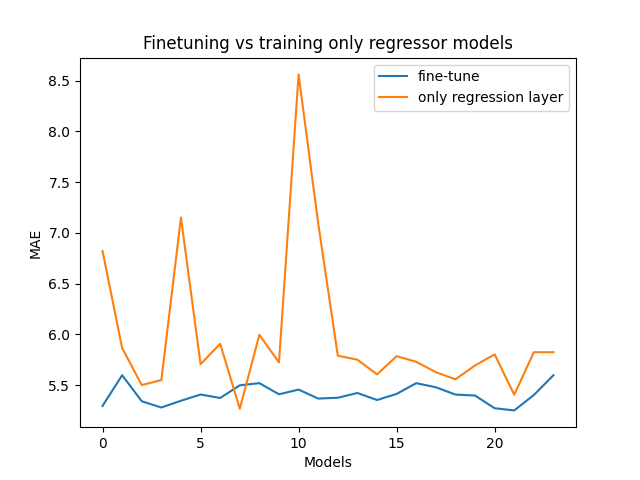}
    \caption{Fine-tuning vs training only regression layer in the models. The x-axis represents the 24 different models of each category. The y-axis shows the MAE corresponding to each model.}
    \label{fig:finetune}
\end{figure}

Fine-tuning on large pretrained language models has become the standard way to conduct NLP research after the widespread adoption of the transformer architecture. And unsurprisingly, our experiments reveal (\cref{fig:finetune}) that fine-tuning of models give smaller MAE scores than training only the regression layers. The stark difference in the variance for the predicted attributes between fine-tuned models and regression only models (as illustrated in \cref{table:10}-\ref{table:11}) further demonstrates the advantage of fine-tuning.

\begin{table}[h!]
\centering
\begin{tabular}{|c|c|c|c|c|} 
\hline
Model & FFD\_$\mu$ & FFD\_$\sigma$ & TRT\_$\mu$ & TRT\_$\sigma$  \\
\hline
Aug  & 5.502 & 2.511 & 8.181 & 5.436 \\
\hline
Uaug  & 5.88 & 2.78 & 9.086 & 6.176\\
\hline
\end{tabular}
\caption{Attribute wise variance of scores for fine-tuned models vs regression-layer only models}
\label{table:10}
\end{table}

\begin{table}[h!]
\centering
\begin{tabular}{|c|c|c|c|c|} 
\hline
Model & FFD\_$\mu$ & FFD\_$\sigma$ & TRT\_$\mu$ & TRT\_$\sigma$ \\
\hline
Aug  & 0.017 & 0.004 & 0.015 & 0.007 \\
\hline
Uaug  & 0.292 & 0.749 & 0.823 & 0.678 \\
\hline
\end{tabular}
\caption{Attribute wise mean of scores for fine-tuned models vs regression-layer only models}
\label{table:11}
\end{table}

\section{Conclusion}
In this paper, we have described our experiments with different kinds of models that were trained on the data provided for this shared-task. We have identified five ways in which we can make better systems to predict eye-tracking features based on eye-tracking data from a multilingual corpus. First, the experiments demonstrate that the inclusion of context (previous words occurring in the sentence) helps the models to predict eye-tracking attributes better. This reaffirms previous observations made with language models that more context is always helpful. Second, we find that XLM based models perform relatively better than the BERT based models. Third, our experiments show the advantages of augmenting additional linguistic features (word length and word frequency information in this case) to the contextual word representations to make better systems. This is in agreement with the findings from eye-tracking prediction tasks from last iterations of CMCL. Fourth, we see how different pooling methods applied on the input token representations affect the final performance of the systems. Finally, the experiments re-validate the approach of fine-tuning pretrained language models for specific tasks. Hence we conclude that contextualized word representations from language models pretrained with many different languages, if carefully augmented, engineered, and fine-tuned, can predict eye-tracking features quite successfully. 


\section{Acknowledgement}
This work has been funded from the grant 19-26934X (NEUREM3) of the Czech Science Foundation.

\bibliography{anthology,custom}

\begin{thebibliography}{22}
\expandafter\ifx\csname natexlab\endcsname\relax\def\natexlab#1{#1}\fi

\bibitem[{Artetxe et~al.(2019)Artetxe, Ruder, and Yogatama}]{artetxe2019cross}
Mikel Artetxe, Sebastian Ruder, and Dani Yogatama. 2019.
\newblock On the cross-lingual transferability of monolingual representations.
\newblock \emph{arXiv preprint arXiv:1910.11856}.

\bibitem[{Choudhary et~al.(2021)Choudhary, Tandon, Agarwal, and
  Chatterjee}]{choudhary2021mtl782_iitd}
Shivani Choudhary, Kushagri Tandon, Raksha Agarwal, and Niladri Chatterjee.
  2021.
\newblock Mtl782\_iitd at cmcl 2021 shared task: Prediction of eye-tracking
  features using bert embeddings and linguistic features.
\newblock In \emph{Proceedings of the Workshop on Cognitive Modeling and
  Computational Linguistics}, pages 114--119.

\bibitem[{Clark et~al.(2019)Clark, Khandelwal, Levy, and
  Manning}]{clark2019does}
Kevin Clark, Urvashi Khandelwal, Omer Levy, and Christopher~D Manning. 2019.
\newblock What does bert look at? an analysis of bert's attention.
\newblock \emph{arXiv preprint arXiv:1906.04341}.

\bibitem[{Conneau et~al.(2020)Conneau, Khandelwal, Goyal, Chaudhary, Wenzek,
  Guzm{\'a}n, Grave, Ott, Zettlemoyer, and Stoyanov}]{conneau2020unsupervised}
Alexis Conneau, Kartikay Khandelwal, Naman Goyal, Vishrav Chaudhary, Guillaume
  Wenzek, Francisco Guzm{\'a}n, {\'E}douard Grave, Myle Ott, Luke Zettlemoyer,
  and Veselin Stoyanov. 2020.
\newblock Unsupervised cross-lingual representation learning at scale.
\newblock In \emph{Proceedings of the 58th Annual Meeting of the Association
  for Computational Linguistics}, pages 8440--8451.

\bibitem[{Dehghani et~al.(2017)Dehghani, Boghrati, Man, Hoover, Gimbel,
  Vaswani, Zevin, Immordino-Yang, Gordon, Damasio
  et~al.}]{dehghani2017decoding}
Morteza Dehghani, Reihane Boghrati, Kingson Man, Joe Hoover, Sarah~I Gimbel,
  Ashish Vaswani, Jason~D Zevin, Mary~Helen Immordino-Yang, Andrew~S Gordon,
  Antonio Damasio, et~al. 2017.
\newblock Decoding the neural representation of story meanings across
  languages.
\newblock \emph{Human brain mapping}, 38(12):6096--6106.

\bibitem[{Devlin et~al.(2019)Devlin, Chang, Lee, and
  Toutanova}]{devlin2019bert}
Jacob Devlin, Ming-Wei Chang, Kenton Lee, and Kristina Toutanova. 2019.
\newblock Bert: Pre-training of deep bidirectional transformers for language
  understanding.
\newblock In \emph{Proceedings of the 2019 Conference of the North American
  Chapter of the Association for Computational Linguistics: Human Language
  Technologies, Volume 1 (Long and Short Papers)}, pages 4171--4186.

\bibitem[{Hollenstein et~al.(2019)Hollenstein, Barrett, Troendle, Bigiolli,
  Langer, and Zhang}]{hollenstein2019advancing}
Nora Hollenstein, Maria Barrett, Marius Troendle, Francesco Bigiolli, Nicolas
  Langer, and Ce~Zhang. 2019.
\newblock Advancing nlp with cognitive language processing signals.
\newblock \emph{arXiv e-prints}, pages arXiv--1904.

\bibitem[{Hollenstein et~al.(2022)Hollenstein, Chersoni, Jacobs, Oseki,
  Pr{\'e}vott, and Santus}]{hollenstein2022shared}
Nora Hollenstein, Emmanuele Chersoni, Cassandra Jacobs, Yohei Oseki, Laurent
  Pr{\'e}vott, and Enrico Santus. 2022.
\newblock Cmcl 2022 shared task on multilingual and crosslingual prediction of
  human reading behavior.

\bibitem[{Hollenstein et~al.(2021)Hollenstein, Pirovano, Zhang, J{\"a}ger, and
  Beinborn}]{hollenstein-etal-2021-multilingual}
Nora Hollenstein, Federico Pirovano, Ce~Zhang, Lena J{\"a}ger, and Lisa
  Beinborn. 2021.
\newblock \href {https://doi.org/10.18653/v1/2021.naacl-main.10} {Multilingual
  language models predict human reading behavior}.
\newblock In \emph{Proceedings of the 2021 Conference of the North American
  Chapter of the Association for Computational Linguistics: Human Language
  Technologies}, pages 106--123, Online. Association for Computational
  Linguistics.

\bibitem[{Khandelwal et~al.(2018)Khandelwal, He, Qi, and
  Jurafsky}]{khandelwal2018sharp}
Urvashi Khandelwal, He~He, Peng Qi, and Dan Jurafsky. 2018.
\newblock Sharp nearby, fuzzy far away: How neural language models use context.
\newblock \emph{arXiv preprint arXiv:1805.04623}.

\bibitem[{Lee et~al.(2019)Lee, Lee, Kim, Kosiorek, Choi, and Teh}]{lee2019set}
Juho Lee, Yoonho Lee, Jungtaek Kim, Adam Kosiorek, Seungjin Choi, and Yee~Whye
  Teh. 2019.
\newblock Set transformer: A framework for attention-based
  permutation-invariant neural networks.
\newblock In \emph{International Conference on Machine Learning}, pages
  3744--3753. PMLR.

\bibitem[{Li and Rudzicz(2021)}]{li2021torontocl}
Bai Li and Frank Rudzicz. 2021.
\newblock Torontocl at cmcl 2021 shared task: Roberta with multi-stage
  fine-tuning for eye-tracking prediction.
\newblock \emph{arXiv preprint arXiv:2104.07244}.

\bibitem[{Liu et~al.(2019)Liu, Gardner, Belinkov, Peters, and
  Smith}]{liu2019linguistic}
Nelson~F Liu, Matt Gardner, Yonatan Belinkov, Matthew~E Peters, and Noah~A
  Smith. 2019.
\newblock Linguistic knowledge and transferability of contextual
  representations.
\newblock In \emph{Proceedings of the 2019 Conference of the North American
  Chapter of the Association for Computational Linguistics: Human Language
  Technologies, Volume 1 (Long and Short Papers)}, pages 1073--1094.

\bibitem[{Mitchell et~al.(2008)Mitchell, Shinkareva, Carlson, Chang, Malave,
  Mason, and Just}]{mitchell2008predicting}
Tom~M Mitchell, Svetlana~V Shinkareva, Andrew Carlson, Kai-Min Chang, Vicente~L
  Malave, Robert~A Mason, and Marcel~Adam Just. 2008.
\newblock Predicting human brain activity associated with the meanings of
  nouns.
\newblock \emph{science}, 320(5880):1191--1195.

\bibitem[{Popel et~al.(2020)Popel, Tomkova, Tomek, Kaiser, Uszkoreit, Bojar,
  and {\v{Z}}abokrtsk{\`y}}]{popel2020transforming}
Martin Popel, Marketa Tomkova, Jakub Tomek, {\L}ukasz Kaiser, Jakob Uszkoreit,
  Ond{\v{r}}ej Bojar, and Zden{\v{e}}k {\v{Z}}abokrtsk{\`y}. 2020.
\newblock Transforming machine translation: a deep learning system reaches news
  translation quality comparable to human professionals.
\newblock \emph{Nature communications}, 11(1):1--15.

\bibitem[{Shao et~al.(2019)Shao, Guo, Chen, and Hao}]{shao2019transformer}
Taihua Shao, Yupu Guo, Honghui Chen, and Zepeng Hao. 2019.
\newblock Transformer-based neural network for answer selection in question
  answering.
\newblock \emph{IEEE Access}, 7:26146--26156.

\bibitem[{Sun et~al.(2019)Sun, Qiu, Xu, and Huang}]{sun2019fine}
Chi Sun, Xipeng Qiu, Yige Xu, and Xuanjing Huang. 2019.
\newblock How to fine-tune bert for text classification?
\newblock In \emph{China national conference on Chinese computational
  linguistics}, pages 194--206. Springer.

\bibitem[{Vaswani et~al.(2017)Vaswani, Shazeer, Parmar, Uszkoreit, Jones,
  Gomez, Kaiser, and Polosukhin}]{vaswani2017attention}
Ashish Vaswani, Noam Shazeer, Niki Parmar, Jakob Uszkoreit, Llion Jones,
  Aidan~N Gomez, {\L}ukasz Kaiser, and Illia Polosukhin. 2017.
\newblock Attention is all you need.
\newblock In \emph{Advances in neural information processing systems}, pages
  5998--6008.

\bibitem[{Vickers et~al.(2021)Vickers, Wainwright, Tayyar~Madabushi, and
  Villavicencio}]{vickers-etal-2021-cognlp}
Peter Vickers, Rosa Wainwright, Harish Tayyar~Madabushi, and Aline
  Villavicencio. 2021.
\newblock \href {https://doi.org/10.18653/v1/2021.cmcl-1.16}
  {{C}og{NLP}-{S}heffield at {CMCL} 2021 shared task: Blending cognitively
  inspired features with transformer-based language models for predicting eye
  tracking patterns}.
\newblock In \emph{Proceedings of the Workshop on Cognitive Modeling and
  Computational Linguistics}, pages 125--133, Online. Association for
  Computational Linguistics.

\bibitem[{Wochna and Juhasz(2013)}]{wochna2013context}
Kacey~L Wochna and Barbara~J Juhasz. 2013.
\newblock Context length and reading novel words: An eye-movement
  investigation.
\newblock \emph{British Journal of Psychology}, 104(3):347--363.

\bibitem[{Wolf et~al.(2020)Wolf, Debut, Sanh, Chaumond, Delangue, Moi, Cistac,
  Rault, Louf, Funtowicz, Davison, Shleifer, von Platen, Ma, Jernite, Plu, Xu,
  Le~Scao, Gugger, Drame, Lhoest, and Rush}]{wolf-etal-2020-transformers}
Thomas Wolf, Lysandre Debut, Victor Sanh, Julien Chaumond, Clement Delangue,
  Anthony Moi, Pierric Cistac, Tim Rault, Remi Louf, Morgan Funtowicz, Joe
  Davison, Sam Shleifer, Patrick von Platen, Clara Ma, Yacine Jernite, Julien
  Plu, Canwen Xu, Teven Le~Scao, Sylvain Gugger, Mariama Drame, Quentin Lhoest,
  and Alexander Rush. 2020.
\newblock \href {https://doi.org/10.18653/v1/2020.emnlp-demos.6} {Transformers:
  State-of-the-art natural language processing}.
\newblock In \emph{Proceedings of the 2020 Conference on Empirical Methods in
  Natural Language Processing: System Demonstrations}, pages 38--45, Online.
  Association for Computational Linguistics.

\bibitem[{Wu and Dredze(2019)}]{wu-dredze-2019-beto}
Shijie Wu and Mark Dredze. 2019.
\newblock \href {https://doi.org/10.18653/v1/D19-1077} {Beto, bentz, becas: The
  surprising cross-lingual effectiveness of {BERT}}.
\newblock In \emph{Proceedings of the 2019 Conference on Empirical Methods in
  Natural Language Processing and the 9th International Joint Conference on
  Natural Language Processing (EMNLP-IJCNLP)}, pages 833--844, Hong Kong,
  China. Association for Computational Linguistics.

\end{thebibliography}
\bibliographystyle{acl_natbib}

\end{document}